\documentclass[journal]{IEEEtran}

\ifCLASSINFOpdf
   \usepackage[pdftex]{graphicx}
   \graphicspath{{./jpeg/}{./authors/}}
   \DeclareGraphicsExtensions{.pdf,.jpeg,.png}
\else

\fi

\usepackage{amsmath,amssymb}

\usepackage{enumitem}
\usepackage{rotating}
\usepackage{url}

\usepackage{breakurl}
\usepackage[breaklinks]{hyperref}

\usepackage{tabularx, booktabs, multirow}
\newcolumntype{C}{>{\centering\arraybackslash}X} 
\begin{document}

\title{Deep Learning-based Vehicle Behaviour Prediction for Autonomous Driving Applications: a Review}

\author{Sajjad~Mozaffari,
        Omar~Y.~Al-Jarrah,
        Mehrdad~Dianati,
        Paul~Jennings,
        
        and~Alexandros~Mouzakitis
\thanks{This  work  was  supported  by  Jaguar  Land  Rover  and  the  U.K.-EPSRC  as part of the jointly funded Towards Autonomy: Smart and Connected Control(TASCC) Programme under Grant EP/N01300X/1}
\thanks{S. Mozaffari, O. Y. Al-Jarrah, M. Dianati, and P. Jennings are with the Warwick Manufacturing Group, University of Warwick, Coventry CV4 7AL, U.K. (e-mail: {sajjad.mozaffari, omar.al-jarrah, m.dianati, Paul.Jennings}@warwick.ac.uk)}
\thanks{A. Mouzakitis is with Jaguar Land Rover Ltd., Coventry CV4 7HS, U.K.}}

\maketitle

\begin{abstract}
 
Behaviour prediction function of an autonomous vehicle predicts the future states of the nearby vehicles based on the current and past observations of the surrounding environment. This helps enhance their awareness of the imminent hazards. However, conventional behaviour prediction solutions are applicable in simple driving scenarios that require short prediction horizons. Most recently, deep learning-based approaches have become popular due to their promising performance in more complex environments compared to the conventional approaches. Motivated by this increased popularity, we provide a comprehensive review of the state-of-the-art of deep learning-based approaches for vehicle behaviour prediction in this paper. We firstly give an overview of the generic problem of vehicle behaviour prediction and discuss its challenges, followed by classification and review of the most recent deep learning-based solutions based on three criteria: input representation, output type, and prediction method. The paper also discusses the performance of several well-known solutions, identifies the research gaps in the literature and outlines potential new research directions.  
\end{abstract}


\IEEEpeerreviewmaketitle

\section{Introduction} \label{intro}

\IEEEPARstart{A}{doption} of autonomous vehicles in the near future is expected to reduce the number of road accidents and improve road safety~\cite{uk-dp-transport}. However, for safe and efficient operation on roads, an autonomous vehicle should not only understand the current state of the nearby road-users, but also proactively anticipate their future behaviour. One part of this general problem is to predict the behaviour of pedestrians (or generally speaking, the vulnerable road-users), which is well-studied in computer vision literature~\cite{socialforce, sociallstm, socialgan, softhardwired}. There are also several review papers on pedestrian behaviour prediction such as ~\cite{human_survey, survey_vision_based, human_survey2}. Another equally important part of the problem is prediction of the intended behaviour of other vehicles on the road. In contrast to pedestrians, vehicles' behaviour is constrained by their higher inertia, driving rules and road geometry, which could help reduce the complexity of the problem, compared to aforementioned problem. Nonetheless, new challenges arise from interdependency among vehicles behaviour, influence of traffic rules and driving environment, and multimodality of vehicles behaviour. Practical limitations in observing the surrounding environment and the required computational resources to execute prediction algorithms also add to the difficulty of the problem, as explained in the later sections of this paper.

There are several published survey papers on vehicle behaviour analysis. For example, Shirazi and Morris~\cite{survey_intersections} provide a review of vehicle monitoring, behaviour and safety analysis at intersections. A review of unsupervised approaches for vehicle behaviour analysis with a focus on trajectory clustering and topic modelling methods is provided in~\cite{survey_unsupervised}. Anomaly detection techniques using visual surveillance are reviewed in~\cite{survey_anomoly}. In~\cite{survey_tracking_pred_dm}
a joint review is provided on tracking, prediction and decision making for autonomous driving. None of these studies specifically focus on vehicle behaviour prediction.
In the most related paper to our work, Lefevre~\textit{et al.}~\cite{survey_motion_risk} provide a survey on vehicle behaviour prediction and risk assessment in the context of autonomous vehicles. The authors review various conventional approaches that applied physics-based models and/or traditional machine learning algorithms such as Hidden Markov Models, Support Vector Machines, and Dynamic Bayesian Networks. Recent advances in machine learning techniques (e.g., deep learning) have provided new and powerful tools for solving the problem of vehicle behaviour prediction. Such approaches have become increasingly important due to their promising performance in complex and realistic scenarios. However, to the best of our knowledge, there is no systematic and comparative review of the latter deep learning-based approaches. We thus present a review of such studies using a new classification method which is based on three criteria: input representation, output type, and prediction method. In addition, we report the practical limitations of implementing recent solutions in autonomous vehicles. To make the paper self-contained, we also provide a generic problem definition for vehicle behaviour prediction.

The rest of this paper is organised to a number of sections: Section \ref{formulation} is an introduction to the basics and the challenges of vehicle behaviour prediction for autonomous vehicles. The definition of used terminologies and the generic problem formulation are also given in section \ref{formulation}. Section \ref{classification} reviews the related deep learning-based solutions and classifies them based on three criteria: input representation, output type, and prediction method. Section \ref{evaluation} discusses the commonly used evaluation metrics, compares the performance of several well-known trajectory prediction models in public highway driving datasets, and highlights the current research gaps in the literature and potential new research directions. The key concluding remarks are given in section \ref{conclusion}.

\section{Basics and Challenges of Vehicle Behaviour Prediction} \label{formulation}

Object detection and behaviour prediction can be considered as two main functions of the perception system of an autonomous vehicle. While both of them rely on on- and off-board sensory data, the former aims to localize and classify the objects in the surrounding environment of the autonomous vehicle and the latter provides an understanding of the dynamics of surrounding objects and predicts their future behaviour. Behaviour prediction plays a pivotal role in autonomous driving applications as it supports efficient decision making~\cite{survey_planning} and enables risk assessment~\cite{survey_motion_risk}. In this section, we firstly discuss the challenges of vehicle behaviour prediction, then we provide a terminology for vehicle behaviour prediction, and finally we present a generic probabilistic formulation of the problem.

\subsection{Challenges}
Vehicles (e.g., cars and trucks) have well-structured motions which are governed by driving rules and environment conditions. In addition, vehicles, as non-holonomic systems, cannot change their trajectories instantly to desired ones. However, vehicle behaviour prediction is not a trivial task due to several challenges. First, there is an interdependency among vehicles behaviour where the behaviour of a vehicle affects the behaviour of other vehicles and vice versa. Therefore, predicting the behaviour of a vehicle requires observing the behaviour of surrounding vehicles. Second, road geometry and traffic rules can reshape the behaviour of vehicles. For example, placing a give-way sign in an intersection can completely change the behaviour of vehicles approaching it. Therefore, without considering traffic rules and road geometry, a model trained in a specific driving environment would have limited performance in other driving environments. Third, the future behaviour of vehicles is multimodal, meaning that given history of motion of a vehicle, there may exist more than one possible future behaviour for it. For example, when a vehicle is slowing down at an intersection without changing its heading direction, both turning right and turning left motions could be expected. A comprehensive behaviour prediction module in an autonomous vehicle should identify all possible future motions to allow the vehicle to act reliably. 

In addition to the intrinsic challenges of the vehicle behaviour prediction problem, implementing a behaviour prediction module in autonomous vehicles comes with several practical limitations. For example, there are restricted computational resources for on-board implementation in autonomous vehicles. In addition, autonomous vehicles can partially observe the surrounding environment using their on-board sensors due to their limitations (e.g., object occlusion, limited sensor range, and sensor noise). Most of existing studies assume having access to a wide unobstructed top-down view of the driving environment which can be obtained by infrastructure sensors (e.g. an infrastructure surveillance camera). Nonetheless, such data can be available if there exists a communication channel with sufficient capacity between the infrastructure and the autonomous vehicle. In addition, it is not cost-effective to cover all road sections with such sensors. Therefore, a behaviour prediction module cannot always rely on unobstructed vision from an infrastructure sensor.

\subsection{Terminology}
To define the problem of vehicle behaviour prediction, we adopt the following terms:  
\begin{itemize}
    \item \textbf{Target Vehicles (TVs)} are the vehicles whose behaviour we are interested in predicting. 
    \item \textbf{Ego Vehicle (EV)} is the autonomous vehicle which observes the surrounding environment to predict the behaviour of TVs. 
    \item \textbf{Surrounding Vehicles (SVs)} are the vehicles whose behaviour is explored by the prediction model as it can potentially impact TV's future behaviour. Different studies may adopt different criteria for selecting SVs based on their modelling assumptions.
    \item \textbf{Non Effective Vehicles (NVs)} are the remaining vehicles in driving environment that are assumed to have no impact on the TV's behaviour.
\end{itemize}

\begin{figure}[!t]
\centering
\includegraphics[width=3.5in]{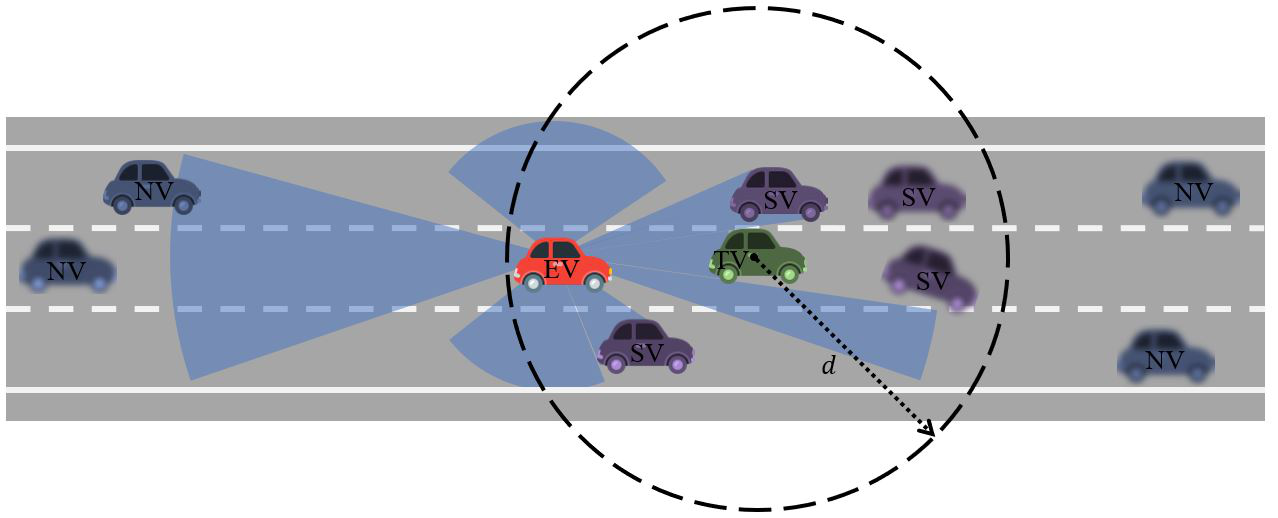}
\caption{An illustration of the adopted terminology and limited observability of the EV's onboard sensors. As an example of a criterion for dividing vehicles into SVs and NVs, the vehicles within a threshold distance, $d$, to the TV are considered to have impact on the TV's behaviour. Unobservable vehicles by the EV, including two of the SVs are represented with blur effect. The limited observability can cause inaccurate prediction. For example, the preceding vehicle of the TV, which is not observable by the EV, is changing its lane allowing the TV to accelerate.}

\label{formulation_fig}
\end{figure}

Figure \ref{formulation_fig} illustrates an example of a driving scenario using the proposed terminology. In this Figure a distance-based criterion, as an example, is used to divide the vehicles into SVs and NVs.

\subsection{Generic Problem Formulation}

We use a probabilistic formulation for vehicle behaviour prediction to cope with the uncertain nature of the problem. The word "behaviour" and "manoeuvre" are sometimes used in the literature interchangeably~\cite{behaviour_man_terms,behaviour_man_terms_1, behaviour_man_terms_2}; however, we consider "vehicle behaviour" as a general term that can imply vehicle manoeuvre or trajectory depending on how it is represented in the problem formulation. In the generic problem formulation, we represent the future behaviour of TVs as the states $X_{TVs}$ they will traverse in the future, defined as:
\begin{equation}
X_{TVs}=\{x_t^i, x_{t+1}^i, ..., x_{t+m}^i \}_{i=1}^N    
\end{equation}
Where $x_t^i$ represents the states (e.g., position) of vehicle $i$ at time step $t$, $N$ is the number of TVs, and $m$ is the length of the prediction window.

The generic problem is formulated as computing the conditional distribution $P(X_{TVs}|O_{EV})$, where $O_{EV}$ are the available observations to the EV. This distribution is a mutual distribution over series of states of several interdependent vehicles, which can be intractable. To reduce the computational requirement of estimating $P(X_{TVs}|O_{EV})$, many of existing works dropped the interdependency among vehicles future behaviour. As such, the behaviour of each TV can be predicted separately with an affordable computational requirement. At each step, one vehicle is selected as the TV and its $P(X_{TV}|O_{EV})$ is calculated, where:
\begin{equation}
X_{TV}=\{x_t^T,x_{t+1}^T, ..., x_{t+m}^T\} 
\end{equation}

Where $T$ is the selected TV.

\section{Classifications of Existing Works} \label{classification}

\begin{figure}[!t]
\centering
\includegraphics[width=3.5in]{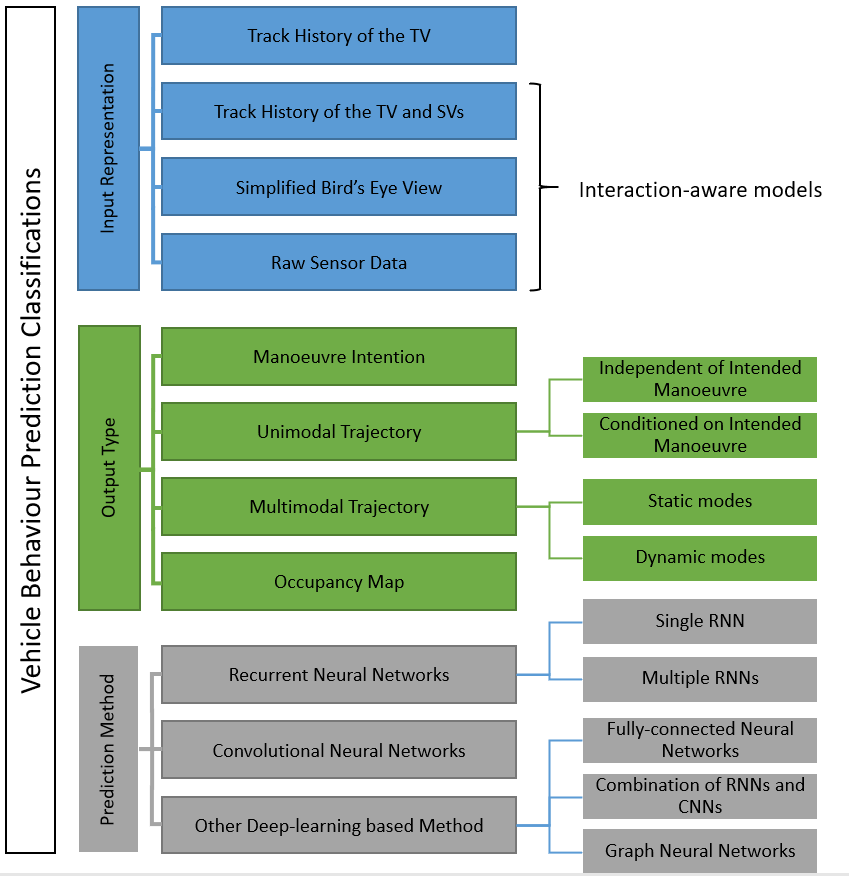}
\caption{
Proposed classifications of state-of-the-art deep learning approaches for vehicle behaviour prediction}

\label{classifications_fig}
\end{figure}

Lefevre~\textit{et al.}~\cite{survey_motion_risk} classifies vehicle behaviour prediction models to physics-based, manoeuvre-based, and interaction-aware models. The simplest approaches that assume the behaviour of vehicles only depends on laws of physics are classified in physics-based models. The models that predict vehicles' behaviour based on their intended manoeuvre are called maneuvre-based approaches. Finally, the more advanced models that consider interaction among vehicles are called interaction-aware models. At the time of writing their paper, in 2014, there has been only a few examples of such interaction-aware model. However, several advanced approaches, mostly deep learning-based has been proposed in the literature since 2014 that requires more detailed classification. Thus, we present three classifications based on three different criteria: input representation, output type, and prediction method. First, we classify existing studies based on how they represent the input data. In this classification, the interaction-aware models are divided into three classes. In second classification, different approaches are classified based on their prediction output. We do not include physics-based approaches as they are no longer state-of-the-art, but different deep learning methods used in behaviour prediction are discussed and classified in the last classification. Figure \ref{classifications_fig} provides the classes and sub-classes for each aforementioned classification criterion. The following subsections address the classification based on each criterion individually.

\subsection{Input Representation}
In this subsection, we provide a classification of existing studies based on the type of input data and how it is represented. We divide them into four classes: track history of the TV, track history of the TV and SVs, simplified bird's eye view, and raw sensor data. The last three classes can be considered as sub-classes of interaction-aware approaches which was introduced in~\cite{survey_motion_risk}. We also discuss the availability of these input data in autonomous driving applications.

\subsubsection{Track history of the TV}
The conventional approach for predicting behaviour of the TV is to only use its current state (e.g. position, velocity, acceleration, heading) or track history of its states over time. This feature can be estimated if the TV is observable by the EV's sensors. 

In~\cite{Sydney_tv_lstm_roundabout, Sydney_tv_lstm_tjunction, sydney_cluster_multi}, the track history of x-y position, speed, and heading of the TV are used to predict its behaviour at different road junctions. All these works study the behaviour of the TV in an environment without any SVs. Few deep learning-based methods use this input set to predict the vehicle behaviour in a driving environment with presence of other vehicles~\cite{dual_lstm, lstm_beam}. Xin \textit{et al.}~\cite{dual_lstm} argue that the information of SVs is not available due to EV's sensor limitations and object occlusion; however, some of the SVs can usually be observed by EV's sensor (see Figure \ref{formulation_fig}). Excluding the observable SV's state from the input set may result in inaccurate prediction of the TV's behaviour due to interdependencies of vehicles' behaviour.

Although the track history of the TV has highly informative features about its short-term future motion, relying only on the TV's track history can lead to erroneous results particularly in long-term prediction in crowded driving environments.

\subsubsection{Track history of the TV and SVs}

One approach to consider the interaction among vehicles is to explicitly feed the track history of the TV and SVs to the prediction model. The SVs' states, similar to the TV's states, can be estimated in the object detection module of the EV; however, some of the SVs can be outside of the EV's sensor range or they might be occluded by other vehicles on the road.

The existing studies vary in how to divide the vehicles in the scene into surrounding vehicles (SVs) and non-effective vehicles (NVs). In~\cite{generalizable, sandiego_lstm_multi, modified_lstm}, history of states of the TV and six of its closest neighbours are exploited to predict the TV's behaviour. In~\cite{semantic, extended_horizon}, the three closest vehicles in the TV's current lane and two adjacent lanes are chosen as reference vehicles. The reference vehicles and the vehicles in front and behind of the two reference vehicles in adjacent lanes are selected as the SVs. The authors in~\cite{lstm_9veh_uni} consider nine vehicles in three lanes surrounding the target vehicle including two vehicles in front of the TV. They indicate that considering more vehicles in the input data can improve the performance of behaviour prediction. For example, in a traffic jam, knowing that the second vehicle ahead of the TV is accelerating can enable early prediction of speed increase for the TV. Instead of considering a fixed number of vehicles as the SVs, a distance threshold is defined to divide vehicle into the SVs and NVs in~\cite{grip, carla, graphCNN_ATT}. It means that only the interactions of vehicles within this threshold are considered in the prediction model. In~\cite{TrafficPredict}, the states of all the observable agents (e.g. vehicles, pedestrian, and cyclist) are used with different weights, obtained by soft attention mechanism~\cite{softAttention}, corresponding to their impacts on TV's behaviour.  

One drawback of most of these  studies is that they assume that the states of all SVs are always observable, which is not a practical assumption in autonomous driving applications. A more realistic approach should always consider sensor impairments like occlusion and noise. In addition, relying only on the track history of the TV and SVs is not sufficient for behaviour prediction, because other factors like environment conditions and traffic rules can also modify the behaviour of vehicles.

\subsubsection{Simplified Bird's Eye View} \label{sbev}

An alternative way to consider the interaction among vehicles is by exploiting a simplified Bird's Eye View (BEV) of the environment. In this approach, static and dynamic objects, road lane, and other elements of the environment are usually depicted with a collection of polygons and lines in a BEV image. The result is a map-like image which preserves the size and location of objects (e.g. vehicles) and the road geometry while ignoring their texture. 
 
Lee~\textit{et al.}~\cite{binaryBEV} fuse front-facing radar and camera data to form a binary two-channel BEV image covering the frontal area of the EV. One of the image channels specifies whether the pixel is occupied by a vehicle or not, and the other depicts the existence of lane marks. For $n$ past frames, the images are produced and stacked together to form a 2n-channel image as the input to the prediction model. Instead of using a sequence of binary images, indicating the existence of objects over time, a single BEV image is used in ~\cite{uber_raster_multi, uber_raster_uni}. In this image, each element of the scene (e.g., road, cross-walks) loses its actual texture and instead is colour coded according to its semantics. The vehicles are depicted by colour-coded bounding boxes and the location history of vehicles are plotted using bounding boxes with same colour and reduced level of brightness.  

To enrich the temporal information within the BEV image, Deo and Trivedi~\cite{sandiego_cnn_multi} use a social tensor which was first introduced in~\cite{sociallstm} (known as social pooling layer). A social tensor is a spatial grid around the target vehicle that the occupied cells are filled with the processed temporal data (e.g., LSTM hidden state value) of the corresponding vehicle. Therefore, a social tensor contains both the temporal dynamic of vehicles and spatial inter-dependencies among them. The authors in~\cite{MATF} add scene context encoding channels to the input representation used in~\cite{sandiego_cnn_multi}. These channels are produced by encoding the static context of the scene (i.e., top-down view image of the scene) using a convolutional neural network. Lee \textit{et al.}~\cite{desire} use social pooling layer as an additional input to another BEV representation created by performing semantic segmentation on front-facing camera of the EV and transforming it to the BEV. 

The aforementioned works do not consider sensor impairment in the input representation. To overcome this drawback, a dynamic occupancy grid map (DOGMa~\cite{dogma}) is exploited in~\cite{dogma_cnn, dogma_rnn}. DOGMa is created from the data fusion of a variety of sensors and provides a BEV image of the environment. The channels of this image contain the probability of occupancy and velocity estimate for each pixel. The velocity information helps distinguish between static and dynamic objects in the environment; however, it does not provide complete knowledge about the history of dynamic objects.

The advantages of simplified BEV is that first it is flexible in terms of complexity of representation. Thus, it can match applications with different computational resource constraints. Second, it enables data fusion from different type of sensors into a single BEV representation. 

One drawback of this input representation, that applies to the previously discussed input representations as well, is that it inherits the limitations of the perception module(e.g., object detection and tracking) used for estimating the states of static and dynamic objects (e.g., vehicles) in the driving environment. Therefore, an error in estimating the states, or under-representing the environment in the perception module will be cascaded to the prediction module. For example, if the object detection module use same label for an ambulance and a normal car, the influence of the ambulance on future behaviour of surrounding vehicles cannot be modelled.

\begin{table*}[]
\centering
\caption{Summary of classification of existing studies based on input representation and the advantages/disadvantages of each class}
\label{table_input}
\begin{tabular}{lllll}
\hline
\textbf{Class}                                                                                  & \textbf{Advantages}                                                                                                                                                                                                                                                                                                                                          & \textbf{Disadvantages}                                                                                                                                                                                                                                    & \textbf{Works}                                                                                                                                                      & \textbf{Summary}                                                                                                                                                                                    \\ \hline
\textbf{\begin{tabular}[c]{@{}l@{}}Track History of\\  the TV\end{tabular}}                     & \begin{tabular}[c]{@{}l@{}}- Complies with limited \\ observability of the EV.\end{tabular}                                                                                                                                                                                                                                                                   & \begin{tabular}[c]{@{}l@{}}- Does not consider the \\ impact of environment and\\ interaction among vehicles\\ on the TV's behaviour.\\ - Inherits the limitation of the\\ perception module of the EV.\end{tabular}                                                  & \begin{tabular}[c]{@{}l@{}}\cite{Sydney_tv_lstm_roundabout, sydney_cluster_multi}, \\  \cite{Sydney_tv_lstm_tjunction, dual_lstm},\\  \cite{lstm_beam}\end{tabular} & \begin{tabular}[c]{@{}l@{}}Track history of the TV's states \\ (e.g., position, velocity, heading, and etc.).\end{tabular}                                                                            \\ \hline
\multirow{5}{*}{\textbf{\begin{tabular}[c]{@{}l@{}}Track History of\\ the TV and SVs\end{tabular}}} & \multirow{5}{*}{\begin{tabular}[c]{@{}l@{}}- Considers the impact of\\  interaction among vehicles on\\ the TV's behaviour.\end{tabular}}                                                                                                                                                                                                                         & \multirow{5}{*}{\begin{tabular}[c]{@{}l@{}}- Does not consider the \\ impact of environment on \\ the TV's behaviour.\\ - The States of SVs are not\\ always observable to the EV.\\ - Inherits the limitation of the\\ perception module of the EV\end{tabular}} & \begin{tabular}[c]{@{}l@{}}\cite{sandiego_lstm_multi, generalizable}, \\  \cite{modified_lstm}\end{tabular}                                                         & History of states for the TV and six SVs.                                                                                                                                                           \\ \cline{4-5} 
                                                                                                &                                                                                                                                                                                                                                                                                                                                                              &                                                                                                                                                                                                                                                           & \cite{semantic, extended_horizon}                                                                                                                                   & \begin{tabular}[c]{@{}l@{}}History of states for the TV and three reference\\ vehicles and four adjacent vehicles to them.\end{tabular}                                                             \\ \cline{4-5} 
                                                                                                &                                                                                                                                                                                                                                                                                                                                                              &                                                                                                                                                                                                                                                           & \cite{lstm_9veh_uni}                                                                                                                                                & History of states for  the TV and nine SVs.                                                                                                                                                         \\ \cline{4-5} 
                                                                                                &                                                                                                                                                                                                                                                                                                                                                              &                                                                                                                                                                                                                                                           & \begin{tabular}[c]{@{}l@{}}\cite{grip, carla}, \\  \cite{graphCNN_ATT}\end{tabular}                                                                                 & \begin{tabular}[c]{@{}l@{}}A distance threshold is defined to divide \\ vehicles into the SVs and NVs.\end{tabular}                                                                                  \\ \cline{4-5} 
                                                                                                &                                                                                                                                                                                                                                                                                                                                                              &                                                                                                                                                                                                                                                           & \cite{TrafficPredict}                                                                                                                                               & \begin{tabular}[c]{@{}l@{}}A soft attention mechanism is used to \\ weight the impact of each observed vehicle.\end{tabular}                                                                         \\ \hline
\multirow{6}{*}{\textbf{\begin{tabular}[c]{@{}l@{}}Simplified Bird's \\ Eye View\end{tabular}}} & \multirow{6}{*}{\begin{tabular}[c]{@{}l@{}}- Considers the impact of\\ environment and interaction \\ among vehicles on the TV's \\ behaviour.\\ - Facilitates fusing the data \\ gathered from different sensors \\ on the EV. \\ - Flexible in terms of\\  complexity of representation.\\ - It can comply with limited \\ observability of the EV.\end{tabular}} & \multirow{6}{*}{\begin{tabular}[c]{@{}l@{}}- Inherits the limitation of the\\ perception module of the EV.\end{tabular}}                                                                                                                                     & \cite{binaryBEV}                                                                                                                                                    & \begin{tabular}[c]{@{}l@{}}A sequence of 2 channel top-down image \\ covering the environment in front of the TV.\\ It indicates the existence of vehicles and\\ lane lines over time.\end{tabular} \\ \cline{4-5} 
                                                                                                &                                                                                                                                                                                                                                                                                                                                                              &                                                                                                                                                                                                                                                           & \cite{uber_raster_multi, uber_raster_uni}                                                                                                                           & \begin{tabular}[c]{@{}l@{}}A BEV image of environment, in which \\ the road elements and vehicles are represented\\  with color-coded polygons and lines.\end{tabular}                              \\ \cline{4-5} 
                                                                                                &                                                                                                                                                                                                                                                                                                                                                              &                                                                                                                                                                                                                                                           & \cite{sandiego_cnn_multi}                                                                                                                                           & \begin{tabular}[c]{@{}l@{}}A top-down grid representation. Each occupied\\ cell is filled with the corresponding vehicle's \\ LSTM hidden state (similar to~\cite{sociallstm}).\end{tabular}       \\ \cline{4-5} 
                                                                                                &                                                                                                                                                                                                                                                                                                                                                              &                                                                                                                                                                                                                                                           & \cite{MATF}                                                                                                                                                         & \begin{tabular}[c]{@{}l@{}}The representation in~\cite{sandiego_cnn_multi} \\ is augmented with CNN encoded image of the \\ static context of the driving scene.\end{tabular}                       \\ \cline{4-5} 
                                                                                                &                                                                                                                                                                                                                                                                                                                                                              &                                                                                                                                                                                                                                                           & \cite{desire}                                                                                                                                                       & \begin{tabular}[c]{@{}l@{}}Semantic segmentation of environment in BEV.\end{tabular}                                                                                                  \\ \cline{4-5} 
                                                                                                &                                                                                                                                                                                                                                                                                                                                                              &                                                                                                                                                                                                                                                           & \cite{dogma_cnn, dogma_rnn}                                                                                                                                         & \begin{tabular}[c]{@{}l@{}}A top-down grid representation. Each cell \\ contains the probability of the cell occupation,\\ and its velocity.\end{tabular}                                           \\ \hline
\multirow{2}{*}{\textbf{\begin{tabular}[c]{@{}l@{}}Raw Sensor \\ Data\end{tabular}}}            & \multirow{2}{*}{\begin{tabular}[c]{@{}l@{}}- Complies with limited \\ observability of the EV.\\ - No information loss.\end{tabular}}                                                                                                                                                                                                                         & \multirow{2}{*}{- High computational cost.}                                                                                                                                                                                                                & \cite{FaF}                                                                                                                                                          & 3D point clouds data over several time steps.                                                                                                                                                      \\ \cline{4-5} 
                                                                                                &                                                                                                                                                                                                                                                                                                                                                              &                                                                                                                                                                                                                                                           & \cite{intentnet}                                                                                                                                                    & \begin{tabular}[c]{@{}l@{}}Lidar data and rasterized map (i.e., the \\ representation used in~\cite{uber_raster_uni}).\end{tabular}                                                                        \\ \hline
\end{tabular}   
\end{table*}

\subsubsection{Raw sensor data}

In this approach, raw sensor data is fed to the prediction model. Thus, the input data contains all available knowledge about the surrounding environment. This allows the model to learn extracting useful features from all available sensory data. 

Raw sensor data, compared to previous input representations, has larger dimension. Therefore, more computational resources are required to process the input data, which can make it impractical for on-board implementation in autonomous vehicles. One solution to this problem is to share the computational resources among different functions of autonomous vehicle. In deep learning literature, it is common to train a model for multiple tasks~\cite{multitasklearning}. In an autonomous vehicle, the object detection module exploits raw sensor data, and it usually relies on a model with millions of parameters~\cite{EA_paper}. Thus, it can be a good candidate for parameter sharing with the behaviour prediction module. 

Leo \textit{et al.}~\cite{FaF} use a deep neural network to jointly solve the problems of 3D detection, tracking, and motion forecasting for autonomous vehicles. They exploit 3D point clouds data over several time frames. The data is represented in BEV images, and the height is considered as the channel dimension. To exploit the lidar data, the same approach is used by~\cite{intentnet}; however, they feed the 3D point cloud data in addition to a simplified BEV to their deep model.

Table \ref{table_input} provides a summary of classification of existing studies based on input representation. It also summarizes the advantages and disadvantages of each class.

\subsection{Output Type}
In this subsection, we classify existing studies based on how they represent a vehicle future behaviour as the output of their prediction model. We consider four classes: manoeuvre intention, unimodal trajectory, multimodal trajectory, and occupancy map.

\subsubsection{Manoeuvre Intention}
Manoeuvre intention prediction (we shortly refer it as intention prediction) is the task of estimating what manoeuvre the vehicle intends to do in upcoming time-steps~\cite{survey_motion_risk}. For example, in highway driving, the set of manoeuvres could be left lane change, right lane change, and keeping the lane; while in an intersection, it could be: go straight, turn left, and turn right. 

To predict the intention of a vehicle approaching a T-junction, Zyner \textit{et al.}~\cite{Sydney_tv_lstm_tjunction} define three classes based on the destination of the vehicle, namely "east", "west", or "south". In~\cite{Sydney_tv_lstm_roundabout}, the same set of classes are used to predict the intention of a vehicle at an un-signalized roundabout. Phillips \textit{et al.}~\cite{generalizable} design a generalizable intention prediction model that can predict the direction of travel of a vehicle up to 150m before reaching three- and four-way intersections. Ding \textit{et al.}~\cite{extended_horizon} and Lee \textit{et al.}~\cite{binaryBEV} apply intention prediction to highway driving scenario. The former proposes an intention prediction model to predict lane change and lane keeping behaviour for the TV; while, the latter designs a model to predict the cut-in intention of right/left preceding TVs w.r.t. the EV. 

Existing studies predict the intention of vehicles using a set of few classes. One	 drawback of these works is that they can only provide a high-level understanding of the vehicle behaviour. This problem can be solved by subdividing high-level manoeuvres into sub-classes that describe the behaviour more precisely. For example, in a highway driving scenario, we can subdivide lane change classes into sharp lane change and normal lane change. Another drawback is the specificity of manoeuvre set to single driving environment, which can be resolved by defining a set that contains the manoeuvres in all desired driving scenarios. However, to predict a vehicle behaviour using large and in depth set of classes, a larger and more diverse training dataset that includes sufficient samples in each class is required. In addition, larger model capacity is needed to learn the mapping of the input data to the intention set.

\subsubsection{Unimodal trajectory}
\begin{figure}[!t]
\centering
\includegraphics[width=3.5in]{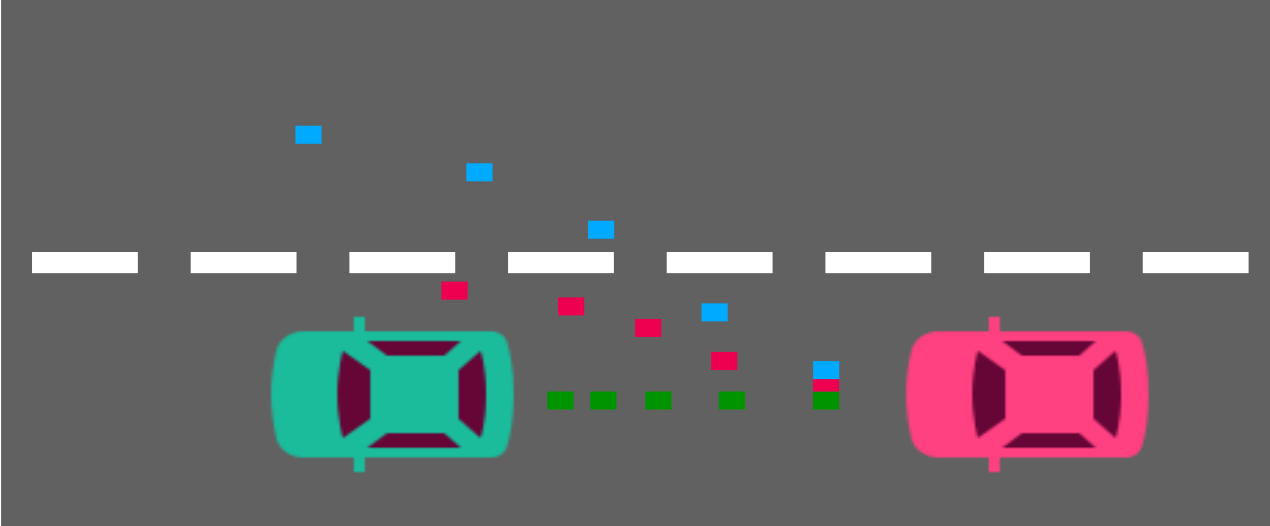}
\caption{An illustration of invalidity of the average of manoeuvres. The red car is approaching the green car on the road. It is probable for the red car to either reduce its speed (green dots) or change its lane (blue dots). A unimodal non manoeuvre-based trajectory predictor may predict an average of these two manoeuvres (red dots) to reduce the prediction error. However, the average of these two manoeuvres is not a valid manoeuvre since it results in a collision with the preceding vehicle. }
\label{mean_man_fig}
\end{figure}

Trajectory prediction models describe the future behaviour of a vehicle by predicting series of future locations of the TV over a time window. Dealing with continuous output of trajectory prediction models can add more complexity to the problem compared to discrete output of intention prediction models. However, predicting trajectory instead of intention, provides more precise information about future behaviour of vehicles. Given a specific driving situation and history of motion for a vehicle, it might be possible for it to traverse multiple different trajectories. Therefore, the corresponding distribution has multiple modes. Unimodal trajectory predictors are the models that only predict one of these possible trajectories (usually the one with highest likelihood). We divide such approaches into two sub-classes:

\begin{itemize}[leftmargin=0cm, itemindent=0.8cm, labelwidth=\itemindent, labelsep=-0.3cm,align=left]

\item \textit{Independent of intended manoeuvre:}
These approaches predict a unimodal trajectory without explicitly considering the effect of possible manoeuvres on it. The straightforward approach to predict the trajectory of the TV is to estimate the position of it over time~\cite{KalmanNN,MATF, grip}. The predictor model can also estimate the displacement of the TV relative to its last position at each step~\cite{graphCNN_ATT, modified_lstm}. The other approach used in~\cite{lstm_9veh_uni} is to predict lateral position and longitudinal velocity separately. This approach can be specially useful when the region of interest is longitudinally large, therefore longitudinal position can be a quite large figure. In addition to the position and velocity, the heading angle of the vehicle is predicted in \cite{FaF}. To cope with uncertainty of the trajectory prediction problem, Djuric \textit{et al.}~\cite{uber_raster_uni} propose a trajectory prediction model that estimates standard deviation for the predicted x- and y-positions. In~\cite{TrafficPredict}, the mean, standard deviation, and correlation coefficient of a bivariate Gaussian distribution corresponding to x- and y- positions are predicted for each time step. The main disadvantage of unimodal trajectory prediction models which are independent of intended manoeuvre is that they may converge to the average of all the possible modes because the average can minimize the displacement error of unimodal trajectory prediction; however, the average of modes is not necessarily a valid future behaviour \cite{sydney_cluster_multi}. Figure \ref{mean_man_fig} illustrates this problem.

\item \textit{Conditioned on intended manoeuvre:} The other unimodal trajectory prediction approaches estimate the likelihood of each member of a predefined manoeuvre intention set and predict the trajectory that corresponds to the most probable intention. Xin \textit{et al.}~\cite{dual_lstm} propose an intention-aware model to predict trajectory based on estimated lane change intention for the TV in highway driving. In~\cite{carla, intentnet}, the intention set is extended from only lane change intentions to turning right, turning left, stopping, and so on. This allows using the prediction model in urban driving. Unimodal trajectory prediction approaches conditioned on intended manoeuvre are unlikely to converge to the mean of modes, as in these approaches the predicted trajectory corresponds to one of predefined behaviour modes. However, there are two main drawbacks in these approaches. First, they cannot accurately predict a vehicle trajectory if the vehicle's intention does not exist in the predefined intention set. This problem can commonly occur in complex driving scenarios, as it is hard to predetermine all possible driving intentions in such environments. Second, unlike previous sub-class, we need to manually label the intention of vehicles in the training dataset, which is time-consuming, expensive and error-prone.
\end{itemize}

\begin{table*}[]
\caption{Summary of classification of existing studies based on output type and the advantages/disadvantages of each class}
\label{table_output}
\begin{tabular}{lllll}
\hline
\textbf{Class}                                                                             & \textbf{Advantages}                                                                                                                                                                                                                                                                                  & \textbf{Disadvantages}                                                                                                                                                                                                                                                                                                                                                                                                                                                      & \textbf{Work}                                                                                               & \textbf{Summary of Output Type}                                                                                                                      \\ \hline
\multirow{4}{*}{\textbf{\begin{tabular}[c]{@{}l@{}}Manoeuvre\\ Intention\end{tabular}}}    & \multirow{4}{*}{\begin{tabular}[c]{@{}l@{}}- Usually has low computational \\ cost.\end{tabular}}                                                                                                                                                                                                     & \multirow{4}{*}{\begin{tabular}[c]{@{}l@{}}- Only provides a high-level \\ understanding of the vehicle behaviour.\\ - Usually covers manoeuvres that are \\ specifically defined for a single driving \\ scenario.\end{tabular}}                                                                                                                                                                                                                                            & \begin{tabular}[c]{@{}l@{}}\cite{Sydney_tv_lstm_tjunction},\\ \cite{Sydney_tv_lstm_roundabout}\end{tabular} & \begin{tabular}[c]{@{}l@{}}The destination of travel at a roundabout and\\  a T-junction.\end{tabular}                                                   \\ \cline{4-5} 
                                                                                           &                                                                                                                                                                                                                                                                                                      &                                                                                                                                                                                                                                                                                                                                                                                                                                                                             & \cite{generalizable}                                                                                        & \begin{tabular}[c]{@{}l@{}}The probabilities of turning right, left, and \\ going straight at an intersection.\end{tabular}                             \\ \cline{4-5} 
                                                                                           &                                                                                                                                                                                                                                                                                                      &                                                                                                                                                                                                                                                                                                                                                                                                                                                                             & \cite{extended_horizon}                                                                                     & \begin{tabular}[c]{@{}l@{}}Lane change behaviour of the TV in \\ highway driving scenarios.\end{tabular}                                              \\ \cline{4-5} 
                                                                                           &                                                                                                                                                                                                                                                                                                      &                                                                                                                                                                                                                                                                                                                                                                                                                                                                             & \cite{binaryBEV}                                                                                            & Right/left cut-in of left/right preceding TVs.                                                                                                       \\ \hline
\multirow{8}{*}{\textbf{\begin{tabular}[c]{@{}l@{}}Unimodal \\ Trajectory\end{tabular}}}   & \multirow{8}{*}{\begin{tabular}[c]{@{}l@{}}General:\\ - Less computational cost \\ compared to multimodal models.\\ Conditioned on intended \\ manoeuvre:\\ - Fixes the problem of convergence \\ to the mean of behaviour modes.\end{tabular}}                                                       & \multirow{8}{*}{\begin{tabular}[c]{@{}l@{}}General:\\ - Does not fully represent the vehicle\\ behaviour prediction space which is \\ multimodal.\\ Independent from intended \\ manoeuvre:\\ - Is prone to convergence to the mean \\ of behaviour modes.\\ Conditioned on intended \\ manoeuvre:\\ - Is prone to trajectory prediction error\\ if the vehicle's intention is not among \\ pre-defined intentions.\\ - Manual labelling is required.\end{tabular}} & \begin{tabular}[c]{@{}l@{}}\cite{KalmanNN}, \\ \cite{grip},\\ \cite{MATF}\end{tabular}                      & \begin{tabular}[c]{@{}l@{}}The positions of the TV over \\ time.\end{tabular}                                                         \\ \cline{4-5} 
                                                                                           &                                                                                                                                                                                                                                                                                                      &                                                                                                                                                                                                                                                                                                                                                                                                                                                                             & \begin{tabular}[c]{@{}l@{}}\cite{graphCNN_ATT}, \\ \cite{modified_lstm}\end{tabular}                        & \begin{tabular}[c]{@{}l@{}}The displacement of the TV relative to its \\ last position for each step.\end{tabular}                                   \\ \cline{4-5} 
                                                                                           &                                                                                                                                                                                                                                                                                                      &                                                                                                                                                                                                                                                                                                                                                                                                                                                                             & \cite{lstm_9veh_uni}                                                                                        & \begin{tabular}[c]{@{}l@{}}Longitudinal velocities and lateral \\ positions over time.\end{tabular}                                                  \\ \cline{4-5} 
                                                                                           &                                                                                                                                                                                                                                                                                                      &                                                                                                                                                                                                                                                                                                                                                                                                                                                                             & \cite{uber_raster_uni}                                                                                      & The x-y position and the standard deviation.                                                                                                         \\ \cline{4-5} 
                                                                                           &                                                                                                                                                                                                                                                                                                      &                                                                                                                                                                                                                                                                                                                                                                                                                                                                             & \cite{TrafficPredict}                                                                                       & \begin{tabular}[c]{@{}l@{}}The mean and variance of a bivariate \\ Gaussian distribution corresponding to\\  x- and y- positions.\end{tabular}        \\ \cline{4-5} 
                                                                                           &                                                                                                                                                                                                                                                                                                      &                                                                                                                                                                                                                                                                                                                                                                                                                                                                             & \cite{FaF}                                                                                                  & \begin{tabular}[c]{@{}l@{}}The bounding box (e.g. location and \\ heading angle) of the TV.\end{tabular}                                             \\ \cline{4-5} 
                                                                                           &                                                                                                                                                                                                                                                                                                      &                                                                                                                                                                                                                                                                                                                                                                                                                                                                             & \cite{dual_lstm}                                                                                            & \begin{tabular}[c]{@{}l@{}}The TV's trajectory (based on lane change \\ estimation for highway driving).\end{tabular}                                \\ \cline{4-5} 
                                                                                           &                                                                                                                                                                                                                                                                                                      &                                                                                                                                                                                                                                                                                                                                                                                                                                                                             & \begin{tabular}[c]{@{}l@{}}\cite{carla}, \\ \cite{intentnet}\end{tabular}                                   & \begin{tabular}[c]{@{}l@{}}The TV's trajectory (based on intention \\ estimation for urban driving).\end{tabular}                                    \\ \hline
\multirow{3}{*}{\textbf{\begin{tabular}[c]{@{}l@{}}Multimodal \\ Trajectory\end{tabular}}} & \multirow{3}{*}{\begin{tabular}[c]{@{}l@{}}General:\\ Potentially can fully represent the \\ vehicle behaviour prediction \\ multimodal space.\\ Dynamic modes: \\ - No manual labelling for behaviour\\ modes is required.\\ - Potentially can adopt to different \\ driving situations.\end{tabular}} & \multirow{3}{*}{\begin{tabular}[c]{@{}l@{}}General:\\ - High computational cost.\\ Dynamic modes:\\ - Is prone to convergence to one \\ behaviour mode or not to explore \\ all the modes.\\ Static modes:\\ Same drawbacks of unimodal \\ models conditioned on intention.\end{tabular}}                                                                                                                                                                               & \begin{tabular}[c]{@{}l@{}}\cite{sandiego_cnn_multi}, \\ \cite{sandiego_lstm_multi}\end{tabular}            & \begin{tabular}[c]{@{}l@{}}The trajectory distribution per each of six\\ predefined manoeuvres and\\  their probability.\end{tabular} \\ \cline{4-5} 
                                                                                           &                                                                                                                                                                                                                                                                                                      &                                                                                                                                                                                                                                                                                                                                                                                                                                                                             & \begin{tabular}[c]{@{}l@{}}\cite{desire},\\ \cite{sydney_cluster_multi},\\ \cite{lstm_beam}\end{tabular}    & \begin{tabular}[c]{@{}l@{}}~\\ A number of samples from the estimated \\ distribution of trajectory.\\ ~\end{tabular}                                \\ \cline{4-5} 
                                                                                           &                                                                                                                                                                                                                                                                                                      &                                                                                                                                                                                                                                                                                                                                                                                                                                                                             & \cite{uber_raster_multi}                                                                                    & \begin{tabular}[c]{@{}l@{}}~\\ A number of deterministic trajectories\\ sequences and their probabilities.\\ ~\end{tabular}                          \\ \hline
\textbf{\begin{tabular}[c]{@{}l@{}}Occupancy\\ Map\end{tabular}}                           & \begin{tabular}[c]{@{}l@{}}- Can potentially predict multiple \\ modes.\end{tabular}                                                                                                                                                                                                                  & \begin{tabular}[c]{@{}l@{}}- prediction accuracy is limited by\\ size of cells of the map.\end{tabular}                                                                                                                                                                                                                                                                                                                                                                     & \begin{tabular}[c]{@{}l@{}}\cite{dogma_cnn}, \\ \cite{dogma_rnn}\end{tabular}                               & \begin{tabular}[c]{@{}l@{}}The probability of occupancy for each \\ pixel of BEV grid map of the driving \\ environment.\end{tabular}                \\ \hline
\end{tabular}
\end{table*}

\subsubsection{Multimodal trajectory}
Multimodal trajectory prediction models predict one trajectory per behaviour modes (a.k.a. policy/manoeuvre/intention) alongside the mode probability. We divide multimodal prediction approaches into two sub-categories:
\begin{itemize}[leftmargin=0cm, itemindent=0.8cm, labelwidth=\itemindent, labelsep=-0.3cm,align=left]
\item \textit{Static modes:} In this sub-class, a set of behaviour modes is explicitly defined and the trajectories are predicted for each member of this set. In~\cite{sandiego_cnn_multi, sandiego_lstm_multi}, a set of six manoeuvre classes for highway driving is defined and the trajectory distribution for each manoeuvre class is predicted. Predicting the distribution allows them to model the uncertainty of trajectory prediction for each manoeuvre separately. Their models also predict the likelihood of each manoeuvre.
\item \textit{Dynamic modes:} In these approaches, the modes can be dynamically learnt based on the driving scenario. Cui \textit{et al.}~\cite{uber_raster_multi} develop a model that predicts a fixed number of deterministic trajectory sequences and their probabilities. Each of these sequences can correspond to a possible manoeuvre in the driving environment. In~\cite{desire, sydney_cluster_multi, lstm_beam}, the distribution of vehicles' trajectory is modelled. Then, a fixed number of trajectory sequences are sampled from the modelled distribution and ranked based on their likelihoods.  
\end{itemize}

The first sub-category of multimodal approaches can be considered as a multimodal extension to unimodal trajectory prediction approaches conditioned on intended manoeuvre as they predict the trajectories for all the behaviour modes rather than the mode with highest likelihood. Therefore, the drawbacks we mentioned for unimodal models conditioned on intended manoeuvre, namely difficulties in defining a comprehensive intention set and manual labelling of intentions in the training dataset, are not solved here. In contrast, the approaches in the second sub-category are exempted from these two problems as they do not require a pre-defined intention set. However, due to dynamic definition of modes, they are prone to converge to a single mode~\cite{sydney_cluster_multi} or not being able to explore all the existing modes.

\subsubsection{Occupancy map}
In these approaches, instead of predicting vehicles trajectories, the occupancy of each cell in a BEV map of the driving environment is estimated for future time-steps. In~\cite{dogma_cnn, dogma_rnn}, the trajectory is predicted by estimating the vehicles occupancy likelihood for each cell in the dynamic occupancy grid map (DOGMa~\cite{dogma}) and each time-step in prediction horizon. They create DOGMa by assigning a grid map to a bird's eye view of the environment around the EV. Their model can dynamically predict multiple trajectory modes by assigning high probability to separate groups of cells in front of a TV. The drawback of such approaches is that their prediction accuracy is limited by the size of the cells in the map. Increasing the number of cells in the grid will reduce the cells' size; however, it results in higher computational costs.

Table \ref{table_output} provides a summary of classification of existing studies based on output type. It also summarizes the advantages and disadvantages of each class. 
\subsection{Prediction Method}
In this subsection, we classify existing studies based on the prediction model used into three classes, namely recurrent neural networks, convolutional neural networks, and other methods.

\begin{table*}[]
\centering
\caption{Summary of classification of existing studies based on the prediction method and the advantages/disadvantages of each class}
\label{table_method}
\begin{tabular}{@{}llll@{}}
\toprule
\textbf{Class}                                                                                       & \textbf{Advantages/Disadvantages}                                                                                                                                                    & \textbf{Work}                                                                                                                         & \textbf{Summary of Prediction Method}                                                                                                                                                                                                                                                                                                                           \\ \midrule
\multirow{10}{*}{\textbf{\begin{tabular}[c]{@{}l@{}}Recurrent\\ Neural\\ Networks\end{tabular}}}     & \multirow{10}{*}{\begin{tabular}[c]{@{}l@{}}- Good at processing \\temporal dependencies.\\Single RNN:\\ - Requires additional \\ mechanism to model \\ interaction and contextual\\ features.\end{tabular}} & \begin{tabular}[c]{@{}l@{}}\cite{Sydney_tv_lstm_tjunction}, \\ \cite{Sydney_tv_lstm_roundabout}, \\ \cite{generalizable}\end{tabular} & Single RNN: Multi-layer LSTM network is used as a sequence classifier.                                                                                                                                                                                                                                                                                          \\ \cmidrule(l){3-4} 
                                                                                                     &                                                                                                                                                                                      & \cite{carla}                                                                                                                          & Single RNN: Two-layer LSTM is used to predict the parameters of acceleration distribution.                                                                                                                                                                                                                                                                      \\ \cmidrule(l){3-4} 
                                                                                                     &                                                                                                                                                                                      & \cite{lstm_9veh_uni}                                                                                                                  & Single RNN: Single-layer LSTM is used to predict future x-y position of  the TV.                                                                                                                                                                                                                                                                                 \\ \cmidrule(l){3-4} 
                                                                                                     &                                                                                                                                                                                      & \cite{lstm_beam}                                                                                                                      & \begin{tabular}[c]{@{}l@{}}Single RNN: An encoder-decoder LSTM is used to predict the probability of the occupancy\\ on a grid BEV.\end{tabular}                                                                                                                                                                                                                 \\ \cmidrule(l){3-4} 
                                                                                                     &                                                                                                                                                                                      & \cite{extended_horizon}                                                                                                               & \begin{tabular}[c]{@{}l@{}}Multiple RNNs: A group of GRUs is used to model the pairwise interaction between the TV \\ and each of the SVs.\end{tabular}                                                                                                                                                                                                          \\ \cmidrule(l){3-4} 
                                                                                                     &                                                                                                                                                                                      & \cite{modified_lstm}                                                                                                                  & \begin{tabular}[c]{@{}l@{}}Multiple RNNs: One group of LSTMs is used to model individual vehicles' trajectory, another\\ group is used to model pairwise interaction.\end{tabular}                                                                                                                                                                               \\ \cmidrule(l){3-4} 
                                                                                                     &                                                                                                                                                                                      & \cite{dual_lstm}                                                                                                                      & \begin{tabular}[c]{@{}l@{}}Multiple RNNs: One LSTM is used to estimate the target lane, another LSTM is used to \\ predict the trajectory based on estimated target lane.\end{tabular}                                                                                                                                                                           \\ \cmidrule(l){3-4} 
                                                                                                     &                                                                                                                                                                                      & \cite{sydney_cluster_multi}                                                                                                           & Multiple RNNs: Multi-layer LSTM are used to predict mixtures of  Gaussian distribution.                                                                                                                                                                                                                                                                         \\ \cmidrule(l){3-4} 
                                                                                                     &                                                                                                                                                                                      & \cite{sandiego_lstm_multi}                                                                                                            & \begin{tabular}[c]{@{}l@{}}Multiple RNNs: One LSTM encoder is applied to the input sequence. The hidden state is fed \\ to six LSTM decoders (one per manoeuvre). Another LSTM encoder is used to predict the\\ probability of each manoeuvre.\end{tabular}                                                                                                      \\ \cmidrule(l){3-4} 
                                                                                                     &                                                                                                                                                                                      & \cite{TrafficPredict}                                                                                                                 & \begin{tabular}[c]{@{}l@{}}Multiple RNNs: multiple LSTMs are grouped as two layers: instance layer and category layer.\\ The former learns instance movement and their interactions, while the latter reason about the\\ similarities of the instance in the same category.\end{tabular}                                                                         \\ \midrule
\multirow{5}{*}{\textbf{\begin{tabular}[c]{@{}l@{}}Convolutional\\ Neural \\ Networks\end{tabular}}} & \multirow{5}{*}{\begin{tabular}[c]{@{}l@{}}- Good at processing\\ spatial dependencies.\\ - 2D CNNs lack a mechanism\\ to  model data series.\end{tabular}}                           & \cite{binaryBEV}                                                                                                                      & \begin{tabular}[c]{@{}l@{}}Six layer CNN with convolution and fully connected layers are used to predict the intention\\  of surrounding vehicles.\end{tabular}                                                                                                                                                                                                 \\ \cmidrule(l){3-4} 
                                                                                                     &                                                                                                                                                                                      & \begin{tabular}[c]{@{}l@{}}\cite{uber_raster_multi},\\ \cite{uber_raster_uni}\end{tabular}                                            & MobileNetV2~\cite{mobilenet} is used as feature extractor.                                                                                                                                                                                                                                                                                                      \\ \cmidrule(l){3-4} 
                                                                                                     &                                                                                                                                                                                      & \cite{dogma_cnn}                                                                                                                      & \begin{tabular}[c]{@{}l@{}}A convolution-deconvolution architecture, introduced in~\cite{deconv}, is used to predict \\vehicle 
                                                                                                 behaviour.
                                                                                                   \end{tabular}                                                                                                                                                                                                    \\ \cmidrule(l){3-4} 
                                                                                                     &                                                                                                                                                                                      & \cite{FaF}                                                                                                                            & \begin{tabular}[c]{@{}l@{}}First, 3D convolutions are applied to the temporal dimension of input data. Then, a series of \\ 2D convolution is used to capture spatial features. Finally, two branches of convolution \\ layers are used to find the probability of being a vehicle and predict the bounding box over \\ current and future frames.\end{tabular} \\ \cmidrule(l){3-4} 
                                                                                                     &                                                                                                                                                                                      & \cite{intentnet}                                                                                                                      & \begin{tabular}[c]{@{}l@{}}First, two backbone CNNs are used to extract the features of lidar data and rasterized map \\ separately. Then three different networks are applied to the concatenation of extracted \\ features to detect vehicles and predict their future intention and trajectory.\end{tabular}                                                                        \\ \midrule
\multirow{7}{*}{\textbf{Other Methods}}                                                              & \begin{tabular}[c]{@{}l@{}}Fully-connected NNs:\\ - Usually rely on current state\\ only.\end{tabular}                                                                                & \cite{semantic}                                                                                                                       & \begin{tabular}[c]{@{}l@{}}~Parameters of vehicle behaviour distribution are estimated using multi-layer fully-connected\\  network.\end{tabular}                                                                                                                                                                                                                \\ \cmidrule(l){2-4} 
                                                                                                     & \multirow{4}{*}{\begin{tabular}[c]{@{}l@{}}Combination of RNNs \\ and CNNs:\\ - Can take advantage of\\ capabilities of both\\ RNNs and CNNs.\end{tabular}}                           & \cite{sandiego_cnn_multi}                                                                                                             & \begin{tabular}[c]{@{}l@{}}An LSTM is applied to each vehicle trajectory. The result is represented in a BEV grid structure\\ and then is fed to a CNN. The output is fed to six LSTM decoders (one per manoeuvre).\end{tabular}                                                                                                                                 \\ \cmidrule(l){3-4} 
                                                                                                     &                                                                                                                                                                                      & \cite{dogma_rnn}                                                                                                                      & \begin{tabular}[c]{@{}l@{}}A convolution network extracts spatial features from the input image. These features are fed to\\ encoder-decoder LSTM. The result is fed to deconvolution network to map to output image with\\  the same size as input.\end{tabular}                                                                                               \\ \cmidrule(l){3-4} 
                                                                                                     &                                                                                                                                                                                      & \cite{desire}                                                                                                                         & \begin{tabular}[c]{@{}l@{}}CVAE-based encoder-decoder GRU generates trajectory distribution. A number of samples from\\ this distribution are ranked and refined based on contextual features.\end{tabular}                                                                                                                                                     \\ \cmidrule(l){3-4} 
                                                                                                     &                                                                                                                                                                                      & \cite{MATF}                                                                                                                           & \begin{tabular}[c]{@{}l@{}}A concatenated vector of agents' movement and static scene encoded by LSTMs and CNNs,\\ respectively are fed to a U-net like network. The encoded movement in the input and output of\\ the mentioned network is fed to LSTM decoders to predict future trajectory for the agents.\end{tabular}                                       \\ \cmidrule(l){2-4} 
                                                                                                     & \multirow{2}{*}{\begin{tabular}[c]{@{}l@{}}Graph Neural Networks:\\ - Comply with graph \\ structure of traffic.\\ - Static scene context is \\ usually neglected.\end{tabular}}       & \cite{graphCNN_ATT}                                                                                                                   & \begin{tabular}[c]{@{}l@{}}Graph Convolutional Network (GCN\cite{graph_cnn}) and Graph Attention Network (GAT\cite{graph_attention}) are \\ used with some adaptations.\end{tabular}                                                                                                                                                                              \\ \cmidrule(l){3-4} 
                                                                                                     &                                                                                                                                                                                      & \cite{grip}                                                                                                                           & \begin{tabular}[c]{@{}l@{}}~\\ Graph Convolutional Model is used which consists of several convolutional and graph operation \\ layers.\end{tabular}                                                                                                                                                                                                            \\ \bottomrule
\end{tabular}
\end{table*}

\subsubsection{Recurrent neural networks}
The simplest recurrent neural network (a.k.a. Vanilla RNN) can be considered as an extension to two-layer fully-connected neural network where the hidden layer has a feedback. This small change allows to model sequential data more efficiently. At each sequence step, the Vanilla RNN processes the input data from current step alongside the memory of past steps, which is carried in the previous hidden neurons. A Vanilla RNN with sufficient number of hidden units can, in principle, learn to approximate any sequence to sequence mapping~\cite{rnn_theory}. However, it is difficult to train this network to learn long sequences in practice due to gradient vanishing or exploding, which is why gated RNNs are introduced~\cite{Graves2012_lstm}. In each cell of these networks, instead of a simple fully connected hidden layer, a gated architecture is deployed. Long short-term memory (LSTM)~\cite{lstm} and Gated recurrent unit (GRU)~\cite{gru} are the most commonly used gated RNNs. In vehicle behaviour prediction, LSTMs are the most used deep models. Here, we sub-categorize recent studies based on the complexity of network architecture:

\begin{itemize}[leftmargin=0cm, itemindent=0.8cm, labelwidth=\itemindent, labelsep=-0.3cm,align=left]
    \item \textit{Single RNN:} In these models, either a single recurrent neural network is used in the simplest form of behaviour prediction (e.g., intention prediction or unimodal trajectory prediction) or a secondary model is used alongside a single RNN to support more sophisticated features like interaction-awareness and/or multimodal prediction. To predict the intention of vehicles, an LSTM is used by~\cite{Sydney_tv_lstm_tjunction, Sydney_tv_lstm_roundabout, generalizable} as a sequence classifier. In this task a sequence of features is fed to successive cells of an LSTM. Then, the hidden state of the last cell in the sequence is mapped to output dimension (i.e., the number of defined classes). In~\cite{Sydney_tv_lstm_tjunction, Sydney_tv_lstm_roundabout}, the input is embedded using a fully-connected layer and is fed to a three-layer LSTM; while, a two-layer LSTM without embedding is used in~\cite{generalizable}. Altché and de La Fortelle~\cite{lstm_9veh_uni} use a single layer LSTM to predict the future x-y position of the TV as a regression task. Despite having less parameters and complexity, single layer LSTMs are reported to achieve competitive results compared to the multilayer counterpart in some tasks~\cite{onhuman,skip}. To predict an intention-based trajectory, Ding and Shen~\cite{carla} use an LSTM encoder to predict the intention of the TV using its states. Then, the predicted intention and map information are used to generate an initial future trajectory for the TV. Finally, a nonlinear optimization method is used to refine the initial future trajectory based on the vehicles interaction, traffic rules (e.g. red lights), and road geometry. To predict multimodal behaviour, Zyner \textit{et al.}~\cite{sydney_cluster_multi} first use an encoder-decoder three-layer LSTM to predict the parameters of a weighted Gaussian Mixture Model (GMM) for each step of the future trajectory. Then, a clustering approach is used to extract the trajectories that correspond to the modes with highest probabilities. Park \textit{et al.}~\cite{lstm_beam} use an encoder decoder LSTM to predict the probability of occupancy on a grid map and apply a beam search algorithm~\cite{beam_search} to select $k$ most probable future trajectory candidates.
    \item \textit{Multiple RNNs:} To deal with multimodality and/or interaction awareness within recurrent neural networks, usually an architecture of several RNNs are used in existing studies. Ding \textit{et al.}~\cite{extended_horizon} use a group of GRU encoders to model the pairwise interaction between the TV and each of SVs, based on which the intention of the TV is predicted for a longer horizon. Dai \textit{et al.}~\cite{modified_lstm} use two groups of LSTM networks for the TV's trajectory prediction, one group for modelling the TV and each of SVs individual trajectory and the other for modelling the interaction between the TV and each of the SVs. Xin \textit{et al.}~\cite{dual_lstm} exploit one LSTM to predict the target lane of the TV and another LSTM to predict the trajectory based on the TV's states and the predicted target lane. To predict multimodal trajectories, the authors in~\cite{sandiego_lstm_multi} use six different decoder LSTMs which correlate with six specific manoeuvres of highway driving. An encoder LSTM is applied to the past trajectory of vehicles. The hidden state of each decoder LSTM is initialized with the concatenation of the last hidden state of the encoder LSTM and a one-hot vector representing the manoeuvre specific to each decoder. The decoder LSTMs predict the parameters of manoeuvre-conditioned bivariate Gaussian distribution of future locations of the TV. Another encoder LSTM is also used to predict the probability of each of six manoeuvres. Multiple LSTMs are structured in~\cite{TrafficPredict} as two main layers, named as instance layer and category layer. The former learns the instance (i.e. agents) movement and their interactions and the latter reason about the similarities among the instances of same category. This network is applied to a graph representation of input data containing 4 dimensions for the instances, their interactions, time, and high-level categorization of instances.
\end{itemize}

Although RNNs are one of the main neural networks associated with data series analysis and prediction such as trajectory prediction, they have deficiency in modelling spatial relationship such as vehicles spatial interaction and image-like data such as driving scene context. This explains why sophisticated solutions using RNNs usually exploit additional methods to compensate the weakness of single RNN.

\subsubsection{Convolutional neural networks}
Convolutional neural networks (CNNs) include convolution layers, where a filter with learnable weights is convolved over the input, pooling layers, which reduce the spatial size of input by sub-sampling, and fully-connected layers, which map their input to desired output dimension.  CNNs are commonly used to extract features from image data. They have achieved successful results in the computer vision domain~\cite{alexnet,vision2}. This success motivates researchers in other domains to represent their data as an image to be able to apply CNNs on them~\cite{conv_seq2seq_human}. However, recently one-dimensional CNNs are also widely used to extract features from one-dimensional signals~\cite{wavenet}.

Lee \textit{et al.}~\cite{binaryBEV} use a six-layer CNN to predict the intention of surrounding vehicles using a binary BEV representation. MobileNetV2~\cite{mobilenet}, which is a memory-efficient CNN designed for mobile applications, is used in~\cite{uber_raster_multi, uber_raster_uni} to extract relevant features from a relatively complex BEV representation. Hoermann \textit{et al.}~\cite{dogma_cnn} use a convolution-deconvolution architecture, which was previously introduced in~\cite{deconv} for image segmentation task, to output the probability of occupancy for future time steps in a BEV image. This model first generates a feature vector using a convolutional network. Then, a deconvolutional network is used to upscale this vector to the output image. A more complex architecture is used in~\cite{intentnet, FaF} to deal with the tasks of object detection and behaviour prediction simultaneously. In~\cite{FaF}, 3D  convolution is performed on the temporal dimension of 4D representation of voxelized lidar data to capture temporal features, then a series of 2D convolutions are applied to extract spatial features. Finally, two branches of convolution layers are added to predict the bounding boxes over the detected objects for current and future frames and estimate the probability of being a vehicle for the detected objects, respectively. In~\cite{intentnet}, two backbone CNNs are used to separately process the BEV lidar input data and the rasterized map. The extracted features are concatenated and fed to three different networks to detect the vehicles, estimate their intention, and predict their trajectories.

Convolutional neural network are valued in vehicle behaviour prediction for their capabilities in taking image-like data, generating image-like output, and keeping spatial relationship of the input data while processing it. These capabilities enables modelling vehicles' interaction and driving scene context and producing occupancy map output. However, 2D CNNs lack a mechanism to model data series which is required in vehicle behaviour prediction for modelling temporal dependencies among vehicles' states over time.

\subsubsection{Other Methods}

\begin{itemize} [leftmargin=0cm, itemindent=0.8cm, labelwidth=\itemindent, labelsep=-0.3cm,align=left]
    
    \item \textit{Fully-connected Neural Networks:} A simplistic approach for vehicle behaviour modelling is to rely only on the current state of the vehicles, which might be inevitable due to unavailability of states history of vehicles or first-order Markov assumption. In this case, the input data is not a sequence and any feed-forward neural networks (e.g. fully-connected neural network) can be used instead of RNNs. In~\cite{mlp}, it is shown that in some driving scenarios, feed-forward neural networks can have competitive results with faster processing time compared to recurrent neural networks. Hu \textit{et al.}~\cite{semantic} use a multi-layer fully connected network to predict the parameters of a Gaussian Mixture Model (GMM). The GMM models the multimodal distribution of arriving time and final location for the TV.
    
    \item \textit{Combination of RNNs and CNNs:} In existing works, recurrent neural networks are used because of their temporal feature extracting power, and convolutional neural networks are used for their spatial feature extracting ability. This inspires some researchers to use both in their models to process both the temporal and spatial dimensions of the data. Nachiket \textit{et al.}~\cite{sandiego_cnn_multi} use one encoder-LSTM per vehicle to extract the temporal dynamics of the vehicle. The internal states of these LSTMs form a social tensor which is fed to a convolutional neural network to learn the spatial interdependencies. Finally, six decoder LSTMs are used to produce the manoeuvre-conditioned distribution of the future trajectory of the TV. In~\cite{dogma_rnn}, a CNN is applied on simplified BEV images each representing the environment around the TV at different time frame. Then, the sequence of extracted features is fed to an Encoder-Decoder LSTM to learn the temporal dynamics of the input data. The decoder LSTM outputs are fed to a deconvolutional neural network to produce output images which represent how the environment around the TV will evolve in the following time steps. In~\cite{desire}, an encoder-decoder GRU is used to generate the distribution of trajectories, then multiple samples of this distribution are fed to decoder GRU to refine and rank them. The latter module also receives the contextual features which are extracted by a CNN model applied on the scene representation. Multi-Agent Tensor Fusion (MATF) encoding and decoding is introduced in~\cite{MATF}. In the encoding part, a social tensor, augmented with convolutional encoded scene context channels, is fed to a U-net~\cite{u_net} like fully convolutional network to fuse interaction among agents and between agents and scene context while keeping spatial locality. Finally, the fused vectors for each vehicle are extracted from the output layer of the U-net like network and are added to the LSTM encoded vectors of the vehicles dynamics and then are fed to LSTM decoders to predict future trajectory per vehicle.
    
    \item \textit{Graph Neural Networks:} The vehicles in a driving scenario and their interaction can be considered as a graph in which the nodes are the vehicles and the edges represent the interaction among them. Using this representation, Graph Neural Networks (GNNs)~\cite{graph_nn1, graph_nn2} can be used to predict TV's behaviour. Diehl \textit{et al.}~\cite{graphCNN_ATT} compare the trajectory prediction performance of two state-of-the-art graph neural networks, namely, Graph Convolutional Network(GCN)~\cite{graph_cnn} and Graph Attention Network (GAT)~\cite{graph_attention}. They also propose some adaptations to improve the performance of these networks for the vehicle behaviour prediction problem. Li \textit{et al.}~\cite{grip} propose a graph-based interaction-aware trajectory prediction (GRIP) model. They use a graph convolutional model, which consists of several convolutional layers as well as graph operations, to model the interaction among the vehicles. The output of the graph convolutional model is fed to an LSTM encoder-decoder to predict the trajectory for multiple TVs. One drawback of current graph-based approach is that static scene context is usually neglected in the modelling procedure.    
\end{itemize}

Table \ref{table_method} provides a summary of classification of existing studies based on the prediction method.

\section{Evaluation} \label{evaluation}
In this section, first we present evaluation metrics that are commonly used for vehicle behaviour prediction in existing studies. Then, the performance of some of existing works is discussed. Finally, we identify and discuss the main research gaps and opportunities.   
\subsection{Evaluation Metrics}
We discuss the evaluation metrics for intention prediction models and trajectory prediction models separately, as the former is a classification problem and the latter is a regression problem and each problem has a separate set of metrics.

\begin{table*}[]
\caption{Comparison of Trajectory Prediction Error of Some of the Existing Works for Different Prediction Horizons}
\label{table_result}
\begin{tabular}{@{}llllccccc@{}}
\toprule
\multicolumn{1}{c}{\multirow{2}{*}{\textbf{Works}}} & \multicolumn{3}{c}{\textbf{Classification}}                                               & \multicolumn{5}{c}{\textbf{RMSE}}                                        \\ \cmidrule(l){2-9} 
\multicolumn{1}{c}{}                                & \textbf{Input Representation} & \textbf{Output Type}       & \textbf{Prediction Model}    & \textbf{1 s} & \textbf{2 s} & \textbf{3 s} & \textbf{4 s} & \textbf{5 s} \\ \midrule
CV                                                  & \multicolumn{1}{c}{-}         & \multicolumn{1}{c}{-}      & \multicolumn{1}{c}{-}        & 0.73         & 1.78         & 3.13         & 4.78         & 6.68         \\ \midrule
~\cite{lstm_9veh_uni}                               & Track history of the TV and SVs   & Unimodal Trajectory        & RNN (Single RNN)             & 0.72         & 2            & 3.76         & 5.97         & 9.01         \\ \midrule
~\cite{dual_lstm}                                   & Track history of the TV           & Unimodal Trajectory & RNN (Multiple RNNs)          & 0.49         & 1.41         & 2.6          & 4.06         & 5.79         \\ \midrule
MATF~\cite{MATF}                                    & Simplified Bird's Eye View    & Unimodal Trajectory        & Combination of RNNs and CNNs & 0.57         & 1.51         & 2.51         & 3.71         & 5.12         \\ \midrule
M-LSTM~\cite{sandiego_lstm_multi}                   & Track history of the TV and SVs   & Multimodal Trajectory      & RNN (Multiple RNNs)          & 0.58         & 1.26         & 2.12         & 3.24         & 4.66         \\ \midrule
CS-LSTM~\cite{sandiego_cnn_multi}                   & Simplified Bird's Eye View    & Multimodal Trajectory      & Combination of RNNs and CNNs & 0.61         & 1.27         & 2.09         & 3.1          & 4.37         \\ \midrule
ST-LSTM~\cite{modified_lstm}                        & Track history of the TV and SVs   & Unimodal Trajectory        & RNN (Multiple RNNs)          & 0.56         & 1.19         & 1.93         & 2.78         & 3.76         \\ \midrule
GRIP~\cite{grip}                                    & Track history of the TV and SVs   & Unimodal Trajectory        & Graph Neural Networks        & 0.64         & 1.13         & 1.8          & 2.62         & 3.6          \\ \bottomrule
\end{tabular}
\end{table*}

\subsubsection{Intention Prediction Metrics}
\begin{itemize}[leftmargin=0cm, itemindent=0.8cm, labelwidth=\itemindent, labelsep=-0.3cm,align=left]
\item \textit{Accuracy:} One of the most common classification metrics is accuracy which is defined as total number of correctly classified data samples divided by total number of data samples. However, relying only on the accuracy can be misleading for an imbalanced dataset. For example, the number of lane changes in a highway driving dataset is usually much less than lane keeping. Thus, an intention predictor that regardless of input data always output lane keeping gains high accuracy score. Therefore, other metrics like precision, recall, and F1 score are also used in existing studies \cite{generalizable, intentnet}.
\item \textit{Precision:} For a given class, precision is defined as the ratio of total number of data samples which are correctly classified in that class to the total number of samples classified as the given class. A low precision indicates a large number of incorrectly classified data as the given class.
\item \textit{Recall:} For a given class, recall is defined as the ratio of total number of data samples which are correctly classified in that class to the total number of samples in the given class. A low Recall indicates a large number of data in the given class that are incorrectly classified in other classes.  
\item \textit{F1 Score:}  The F1 score (a.k.a. F-score or F-measure) is a balance between precision and recall and is defined as:
\begin{equation}
    F_1 = 2\cdot \frac{precision\cdot recall}{precision+recall}
\end{equation}
\item \textit{Negative Log Likelihood (NLL):} For each data sample in a multi-class classification task, NLL is calculated as:
    
\begin{equation}
    NLL = -\sum_{c=1}^{M}y_clog(\hat{y}_c)
\end{equation}
Where $y_c$ is a binary indicator of correctness of predicting the data sample in class $c$, $\hat{y}_c$ is the predicted probability of the data sample belonging to class $c$, and $M$ is the number of classes. Although NLL values are not as interpretable as previously discussed metrics, it can be used to compare the uncertainty of different intention prediction models~\cite{extended_horizon}. 

\item \textit{Average Prediction Time:} This metric is used in intention prediction approaches~\cite{extended_horizon, semantic}, such as lane change prediction, where the approach is applied on a sliding window of the input data series to predict the occurrence of a positive class(e.g., lane change). The metric is obtained by taking the average of the time of the first correct positive class prediction for all samples, considering the time of lane change occurrence as the origin. In~\cite{extended_horizon}, they considered the time when a consistent correct lane change prediction starts to increase robustness of the metric. 

\end{itemize}

\subsubsection{Trajectory Prediction Metrics}
The following metrics are the commonly used metrics in the literature. A detailed discussion on other trajectory prediction metrics can be found in~\cite{traj_metrics}.
\begin{itemize} 
[leftmargin=0cm, itemindent=0.8cm, labelwidth=\itemindent, labelsep=-0.3cm,align=left]

\item \textit{Final Displacement Error (FDE):} This error measures the distance between predicted final location $\hat{y}_{t_{final}}$ and true final location of the TV $y_{t_{final}}$ at the end of prediction horizon $t_{final}$ , while it does not consider the prediction error occurred in other time steps in the prediction horizon. 
    
    \begin{equation}
          FDE = |\hat{y}_{t_{final}} - y_{t_{final}}| 
    \end{equation}

\item \textit{Mean Absolute Error (MAE) or Root Mean Squared Error (RMSE):} MAE measures the average magnitude of prediction error $e_t$, while RMSE measures the square root of the average of the squared prediction error:
    
    \begin{equation}
         MAE = \frac{1}{n}\sum_{t=1}^{n}|e_t|
    \end{equation}
    
    \begin{equation}
         RMSE = \sqrt{\frac{1}{n}\sum_{t=1}^{n}e_t^2}
    \end{equation}
    
    Where $n$ is number of data samples and $e_t$ can be defined as the displacement error between the predicted trajectory and the ground truth. MAE and RMSE are two of the most common metrics for regression problems and act roughly similar. However, RMSE is more sensitive to large errors due to usage of squared error in its definition.

\item \textit{Minimum of K Metric}: In some of existing multimodal trajectory prediction studies~\cite{socialgan, uber_raster_multi, desire, r2p2}, where $K$ trajectories are predicted for different modes, the metric (e.g., MSE, FDE) is calculated using one of the $K$ trajectories that minimize the metric (i.e., best predicted trajectory). The main shortcoming of this evaluation method, also discussed in~\cite{r2p2}, is that the quality of ignored $K-1$ trajectories is not examined. Therefore, a model, reported to have high performance using this metric, can have mostly poor predictions.

\item \textit{Cross Entropy:} For a modelled trajectory distribution $q$, and ground truth data distribution $p$, the cross entropy can be calculated as:
    \begin{equation}
    H(p,q) = \mathop{\mathbb{E}}_{x\sim p}-log(q(x))
    \end{equation}
    Cross entropy (a.k.a. Negative Log Likelihood) can be reported as a metric in both intention prediction and trajectory prediction; however, in multimodal trajectory prediction this metric can be more important as both MAE and RMSE are biased in favour of models that predict the average of modes~\cite{sandiego_cnn_multi} which is not necessarily a good prediction, as discussed before. Although cross entropy penalises a multimodal prediction model for not covering all the modes of ground truth data distribution, it will assign relatively low penalty for a model that predict other modes in addition to ground truth modes. Therefore, Rhinehart~\textit{et al.}~\cite{r2p2} propose using a symmetrized cross entropy metric which is defined as:
   \begin{equation}
    H(p,q) + H(q,\bar{p}) = \mathop{\mathbb{E}}_{x\sim p}-log(q(x)) + \mathop{\mathbb{E}}_{x\sim q}-log(\bar{p}(x))
    \end{equation} 
where $\bar{p}$ is an approximate to $p$, as it is not possible to evaluate the ground truth data distribution $p'$s PDF.     
  
\item \textit{Computation Time:} The trajectory prediction models are usually more complex compared to intention prediction models. Therefore, they can take more computation time which might make them impractical for on-board implementation in autonomous vehicles. Thus, it is crucial to report and compare computation time in trajectory prediction models.
\end{itemize}

\subsection{Performance of Existing Methods}
In this part we compare the performance of some of reviewed trajectory prediction methods. The selected studies for comparison are the ones that used common publicly available datasets and common metrics. These studies report RMSE errors for prediction horizons of 1.0 to 5.0s on NGSIM I-80 and US-101 highway driving datasets~\cite{NGSIM}. Table \ref{table_result} provides the reported error for each model which is obtained from the original paper (except the RMSE calculation of~\cite{lstm_9veh_uni} which has been modified by~\cite{dual_lstm} to match the position error in SI units). Note that the RMSE error is reported for longitudinal and lateral position separately in~\cite{dual_lstm}; however, we calculated the total RMSE error to be consistent with other studies. Furthermore, in~\cite{modified_lstm} the error is calculated for US-101 and I-80 separately; while, we report the average of them. We also report the prediction result of a constant velocity Kalman Filter(CV) model as a simple baseline which is obtained from~\cite{grip}.

To compare the performance of selected works, Table~\ref{table_result} states the category each work belongs to. According to the table \ref{table_result}, most of deep learning-based methods surpass the simple baseline constant velocity model (CV) with a high margin. Among reviewed deep learning-based models, complex models (e.g., Multiple RNNs or Combination of RNNs and CNNs) achieve better performance compared to simple models like single RNN. Nonetheless, increasing the complexity of output, by predicting multimodal trajectory instead of unimodal trajectory, does not always result in lower RMSE. For example, the models named GRIP~\cite{grip} and ST-LSTM~\cite{modified_lstm} achieve better performance compared to M-LSTM~\cite{sandiego_lstm_multi} and CS-LSTM~\cite{sandiego_cnn_multi}, while the former studies predict unimodal trajectories and the latter ones predict multimodal trajectories. This can be due to limited model capacity or limited data used in training the discussed multimodal trajectory prediction models.

\subsection{Research Gaps and New Opportunities} 
\begin{figure}[!t]
\centering
\includegraphics[width=3.5in]{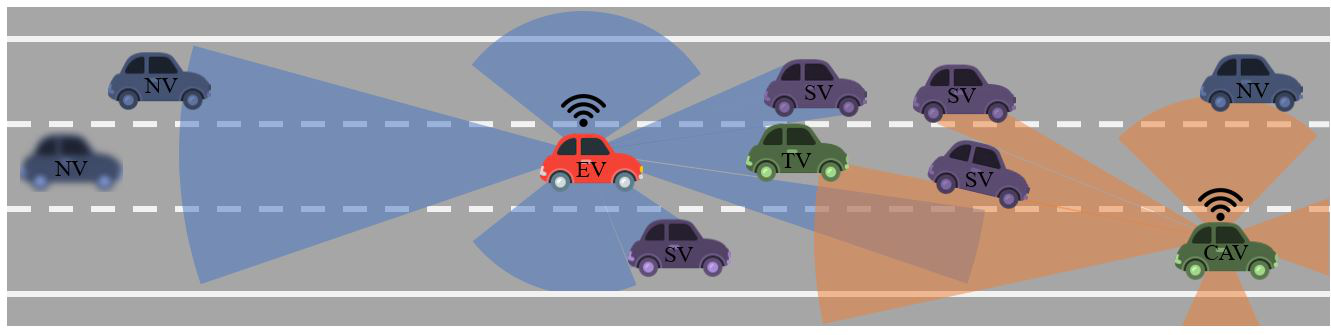}
\caption{An illustration of the vehicle behaviour prediction problem for connected autonomous vehicles. The sensors implemented in other autonomous vehicles and infrastructure can provide more information about the SVs and reduce the object occlusion problem in ego vehicle. }
\label{cav}
\end{figure}

We discuss some of the main research gaps in vehicle behaviour prediction problem, which can be considered as opportunities for future works:
\begin{enumerate} [leftmargin=0cm, itemindent=0.8cm, labelwidth=\itemindent, labelsep=-0.3cm,align=left]
  \item Unlike object detection which has unified way of evaluation~\cite{EA_paper}, there is no benchmark for evaluating existing studies on vehicle behaviour prediction. This prevents a fair comparison among different deep learning-based approaches and between deep learning-based and other methods. For example, among the reviewed deep learning-based papers, there are only seven works that use unified evaluation method (the works that we compare their performance in this paper). In addition, only a few works report the computation time of their algorithms, while this metric is highly important in autonomous driving applications. As a future work, a benchmark can be defined and used in vehicle behaviour prediction to be able to thoroughly compare the performance of different studies.

  \item Most of the existing works consider full observability of the surrounding environment and vehicles' states which is not feasible in practice. Infrastructure sensors can provide non-occluded top-down view of the environment; however, it is impractical to cover all road sections with such sensors. Therefore, a realistic solution for behaviour prediction should always consider sensor impairments (e.g. occlusion, noise) which can limit the number of observable vehicles around the TV and in turn may reduce the accuracy of behaviour predictors in autonomous vehicles. One possible solution is the utilization of connected autonomous vehicles. In this case, the connected vehicle can exploit the information gained by sensors implemented in other vehicles or infrastructure through V2V and V2I communication (see Figure \ref{cav}).
  
  \item In recent studies, traffic rules are rarely considered as an explicit input to the model; while, they can reshape the behaviour of a vehicle in a driving scenario. Some of the existing studies include road direction or traffic light as an input to the prediction model \cite{generalizable, carla} which are only a small part of traffic signs and rules. 
  
  \item In addition to the vehicle's states and scene information which both are usually considered in recent works, other visual and auditory data of vehicles, like vehicle's signalling lights and vehicle horn can also be used to infer about its future behaviour.
  
  \item Most of the existing works are limited to a specific driving scenario such as roundabout, intersection, and T-junction. However, a vehicle behaviour prediction module in fully autonomous vehicle should be able to predict the behaviour in any driving scenario. Developing a model which can be applied to a variety of driving environment can be a direction for future research.
  
\end{enumerate}

\section{Conclusion} \label{conclusion}
Although deep learning-based behaviour prediction solutions have shown promising performance, especially in complex driving scenarios, by utilizing sophisticated input representation and output type, there are several open challenges that need to be addressed to enable their adoption in autonomous driving applications. Particularly, while most of existing solutions considered the interaction among vehicles, factors such as environment conditions and set of traffic rules are not directly inputted to the prediction model. In addition, practical limitations such as sensor impairments and limited computational resources have not been fully taken into account.

\ifCLASSOPTIONcaptionsoff
  \newpage
\fi

\bibliographystyle{IEEEtran}

\bibliography{references}

\begin{IEEEbiography}
[
{
\includegraphics[width=1in,height=1.25in,clip,keepaspectratio]{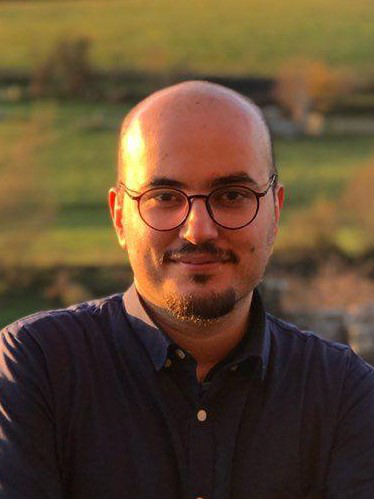}
}
]
{Sajjad Mozaffari}
is a PhD candidate with the Warwick Manufacturing Group (WMG) at University of Warwick, UK. He received the B.Sc. and M.Sc. degrees in Electrical Engineering at the University of Tehran, Iran in 2015 and 2018, respectively. His research interests include machine learning, computer vision, and connected and autonomous vehicles.
\end{IEEEbiography}

\begin{IEEEbiography}
[
{
\includegraphics[width=1in,height=1.25in,clip,keepaspectratio]{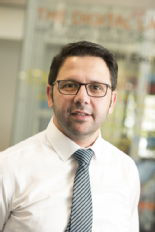}
}
]
{Omar Y. Al-Jarrah}
received the B.S. degree in Computer Engineering from Yarmouk University,  Jordan, in 2005, the MSc degree in Engineering from The University of Sydney,  Sydney, Australia in 2008 and the Ph.D.degree in Electrical and Computer Engineering from Khalifa University, UAE, in 2016. Omar has worked as a postdoctoral fellow in the Department of Electrical and Computer Engineering, Khalifa University, UAE, and currently he works as a senior research fellow at WMG, The University of Warwick, U.K. His main research interest involves machine learning, connected and autonomous vehicles, intrusion detection, big data analytics, and  knowledge discovery in various applications. He has authored/co-authored several publications on these topics. Omar has served as TPC member of several conferences, such as IEEE Globecom 2018. He was the recipient of several scholarships during his undergraduate and graduate studies.
\end{IEEEbiography}

\begin{IEEEbiography}
[
{
\includegraphics[width=1in,height=1.25in,clip,keepaspectratio]{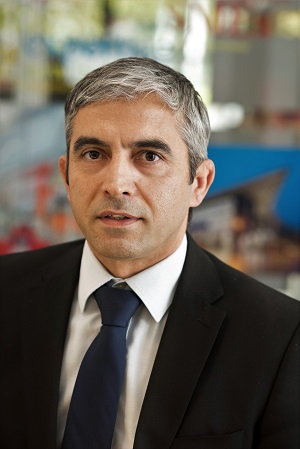}
}
]
{Mehrdad Dianati}
is a Professor of Autonomous and Connected Vehicles at Warwick Manufacturing Group (WMG), University of Warwick, as well as, a visiting professor at 5G Innovation Centre (5GIC), University of Surrey, where he was previously a Professor. He has been involved in a number of national and international  projects as the project leader and work-package leader in recent years. Prior to his academic endeavour, he have worked in the industry for more than 9 years as senior software/hardware developer and Director of R\&D. He frequently provide voluntary services to the research community in various editorial roles; for example, he has served as an associate editor for the IEEE Transactions on  Vehicular Technology, IET Communications and Wiley’s Journal of Wireless Communications and Mobile.
\end{IEEEbiography}

\begin{IEEEbiography}
[
{
\includegraphics[width=1in,height=1.25in,clip,keepaspectratio]{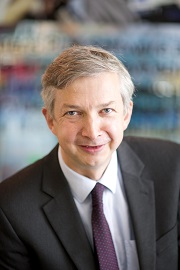}
}
]
{Paul Jennings}
received a BA degree in physics from the University of Oxford in 1985 and an Engineering Doctorate from the University of Warwick in 1996. Since 1988 he has worked on industry-focused research for WMG at the University of Warwick. His current interests include: vehicle electrification, in particular energy management and storage; connected and autonomous vehicles, in particular the evaluation of their dependability; and user engagement in product and environment design, with a particular focus on automotive applications.
\end{IEEEbiography}

\begin{IEEEbiography}
[
{
\includegraphics[width=1in,height=1.25in,clip,keepaspectratio]{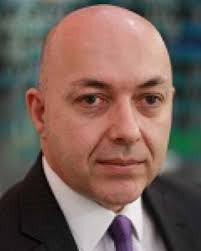}
}
]
{Alexandros Mouzakitis}
is the head of the Electrical, Electronics and Software Engineering Research Department at Jaguar Land Rover. Dr Mouzakitis has over 15 years of technological and managerial experience especially in the area of automotive embedded systems. In his current role is responsible for leading a multidisciplinary research and technology department dedicated to deliver a portfolio of advanced research projects in the areas of human-machine interface, digital transformation, self-learning vehicle, smart/connected systems and onboard/off board data platforms. In his previous position within Jaguar Land Rover, Dr Mouzakitis served as the head of the Model-based Product Engineering department responsible for model-based  development and automated testing standards and processes.

\end{IEEEbiography}

\end{document}